# SIMPL: Generating Synthetic Overhead Imagery to Address Zero-shot and Few-Shot Detection Problems

Yang Xu, Bohao Huang, Xiong Luo, Kyle Bradbury, and Jordan M. Malof

**Abstract**
**Recently deep neural networks (DNNs) have achieved tremendous success for object detection in overhead (e.g., satellite) imagery. One ongoing challenge however is the acquisition of training data, due to high costs of obtaining satellite imagery and annotating objects in it. In this work we present a simple approach – termed Synthetic object IMPLantation (SIMPL) - to easily and rapidly generate large quantities of synthetic overhead training data for custom target objects. We demonstrate the effectiveness of using SIMPL synthetic imagery for training DNNs in zero-shot scenarios where no real imagery is available; and few-shot learning scenarios, where limited real-world imagery is available. We also conduct experiments to study the sensitivity of SIMPL's effectiveness to some key design parameters, providing users for insights when designing synthetic imagery for custom objects. We release a software implementation of our SIMPL approach so that others can build upon it, or use it for their own custom problems.**

## I. INTRODUCTION

Deep neural networks (DNNs) now dominate benchmark problems for object detection in overhead imagery (e.g., satellite imagery or aerial photography). DNNs are high-capacity non-linear models that must be trained to perform object detection using large quantities of annotated imagery (e.g., annotated with bounding boxes). Therefore a crucial contributor to this recent success has been the development of large publicly-available benchmark datasets of overhead imagery that can be employed for DNN training. Some notable recent benchmark datasets include xView[1], DOTA[2] and functional map of the world (FoW)[3].

While these datasets have been crucial to enable DNN research, the performance of DNNs on these datasets represents their recognition capabilities under relatively idealistic conditions. Specifically, these datasets contain hundreds of square kilometers of overhead imagery, and hundreds or thousands of manually annotated target objects (e.g., cars, buildings, aircraft etc.). [1] By contrast, real-world applications often involve novel classes of target objects for which there may be little or no labeled data available. In these cases it is necessary to collect training imagery for the new target objects.

The collection of training imagery is time-consuming and costly in many application domains, but these difficulties are even more severe for applications involving overhead imagery. This is primarily a consequence of the high costs of procuring overhead imagery (especially high resolution imagery), *and* the potentiality that few or no target instances may exist in any particular collection of purchased imagery. As a result, one can purchase and manually inspect a large collection of imagery without finding enough instances of the target object to effectively train a DNN. The number of instances needed to achieve satisfying recognition results also varies based on many factors (e.g., the recognition task, or target object). Therefore the process of procuring training data in overhead imagery is not only costly, but also risky since the investment required to collect a satisfactory dataset are not precisely known in advance.

Notably, these challenges also grow in proportion to the *infrequency* of the target object. As the probability of finding the object gets smaller, more overhead imagery must be (i) purchased and (ii) inspected to find each unique instance of the target object. Therefore, as researchers and practitioners build recognition systems that involve more unique or task-specific target objects, the challenges with collecting training imagery will also grow rapidly.

One way to address these problems is to develop DNN models that can recognize new classes of target objects with very few, or even zero, instances of training data. This problem is known as *N*-shot learning, where *N* refers to the

This work was supported by the China Scholarship Council, in part by the National Natural Science Foundation of China under Grant U1836106, in part by the Beijing Natural Science Foundation under Grants 19L2029 and M21032, Beijing Intelligent Logistics System Collaborative Innovation Center under Grant BILSCIC-2019KF-08, in part by the Scientific and Technological Innovation Foundation of Shunde Graduate School, USTB, under Grant BK20BF010, and in part by the Fundamental Research Funds for the University of Science, and Technology Beijing under Grant FRF-BD-19-012 A.

Yang Xu and Xiong Luo are with the School of Computer and Communication Engineering, University of Science and Technology Beijing, Beijing 100083, China, and with the Beijing Key Laboratory of Knowledge Engineering for Materials Science, Beijing 100083, China, and also with the Shunde Graduate School, University of Science and Technology Beijing, Foshan 528399, China (e-mail: b20160304@xs.ustb.edu.cn; xluo@ustb.edu.cn).

Bohao Huang, Kyle Bradbury, and Jordan M. Malof are with the Electrical and Computer Engineering Department at Duke University, Durham, NC 27705, U.S.A. (email: bohao.huang@duke.edu; kyle.bradbury@duke.edu; jmmalo03@gmail.com). Kyle Bradbury is also with the Energy Initiative at Duke University.

number of real-world training instances available for each new class of target objects [4], [5]. Typically $N$ is assumed to be very small or even zero, in which case it is often called few-shot or zero-shot learning, respectively. There has been substantial work on $N$-shot learning over the past decade, resulting in many advances (e.g., [4]–[7] ).

In general, however, there is still the need to collect and label *some* real-world training imagery in order to achieve reliable recognition accuracy with DNNs, and accuracy can usually be improved substantially with larger quantities of training data. Further, and as discussed, the collection of training data is especially time-consuming, costly and risky in the context of overhead imagery. As a result, extending DNNs to new classes of objects in a reliable and cost-effective manner is a major obstacle to the widespread use of DNN-based recognition in overhead imagery.

In this work we explore the use of *synthetic* overhead imagery to train object detection models. Here "synthetic imagery" refers to imagery that has been captured from a simulated camera operating over a virtual world. In a virtual world, the designer can create large numbers of target objects at little additional cost, and position them within different contexts to create visual diversity (e.g., different lighting conditions, and background content). Furthermore, because the target locations are known by design, precise and accurate ground truth information is readily obtained. With these capabilities, synthetic imagery may permit the training of recognition models with very little, or no, real-world training, thereby addressing $N$-shot learning problems for overhead imagery.

*A. Related work*

In the last decade researchers have found substantial success utilizing synthetic imagery to help train DNNs to solve a variety of problems [8]–[16]. Some notable examples were presented in [8] (the SYNTHIA dataset) and [9], in which synthetic imagery with pixel-wise semantic labels were generated from 3D virtual worlds. It has been shown that these synthetic datasets can boost the performance of segmentation networks on real-world benchmark imagery, such as for this task, such as the CamVid [17] and KITTI [18] dataset, among others.

*Limitations of existing tools for generating synthetic overhead imagery.* While substantial work has been done on synthetic imagery, relatively little work has been done specifically with synthetic *overhead* imagery – our goal here. One major reason for this, as discussed in [19], is that conventional resources for creating synthetic imagery are not well-suited for the creation of *overhead* imagery. For example, existing tools focus on achieving high photorealism for *ground-level* imagery, which requires photorealism at a much finer spatial scale than overhead imagery. One consequence of this is that substantial design effort is required to create even a relatively small virtual world using conventional tools, making them inefficient (or altogether intractable) for designing the large virtual worlds (e.g., tens or hundreds of square kilometers) that are needed for synthetic *overhead* imagery.

*Training DNNs with synthetic overhead imagery.* Perhaps the first work exploring the use of synthetic *overhead* imagery was in 2017 [20], where the authors used several private software tools to generate patches of synthetic overhead imagery that each contained a target object. The authors presented quantitative experimental results for one class of target objects (a helicopter) and found that augmenting real-world training imagery with their synthetic patches resulted in substantial improvement in the accuracy of their DNN-based detection model.

In more recent work [21] the authors developed a large dataset of synthetic and real-world overhead imagery, termed RarePlanes, which includes box labels and attributes for a variety of aircraft. The authors found that they could substantially reduce the amount of real-world overhead imagery needed to train their DNN-based detectors by augmenting with their synthetic imagery(e.g., by roughly 90%).

One general challenge when using synthetic imagery is the visual "domain gap" [11], [12], [19], which refers to the visual differences that exist between synthetic and real-world imagery, even when the content in both images is similar. The recent work in [22] trained a deep learning model to stylize synthetic overhead imagery to be more visually similar to real-world overhead imagery, and found that this improved the effectiveness of their synthetic imagery. To test their approach they created their own synthetic imagery by overlaying 3D models of target objects onto real-world ovheread imagery. This approach is similar to the one we propose here, however there are several key differences (outlined next), and our approach was developed independently and concurrently.

Although these existing works make valuable contributions, their methods are difficult to replicate or build upon – a cornerstone of research. One reason for this is that none of these works share the details of their synthetic generation process, nor do they share software to implement their process, or vary the designs of the imagery for new problems. Another major limitation is that these approaches do not fully describe the design details of their synthetic imagery (e.g., scene content, textures, sizes, locations), or how they chose these parameters. To deploy these techniques in practice, we ultimately need a clear and efficient process to design the synthetic imagery.

*Towards a an more replicable and practical process for synthetic overhead imagery generation.* To our knowledge, the work in [19] (2020) is the only one to publish accessible software for creating their synthetic imagery, as well as full details about the design of their synthetic imagery. As a result, our work here builds primarily upon their methodology, and we too release our software and synthetic data design details. The authors in [19] employed their synthetic generation method to create Synthinel-1, a large dataset of synthetic overhead imagery for the task of building segmentation. The authors presented experiments demonstrating that Synthinel-1 is effective for augmenting real-world imagery when training DNNs.

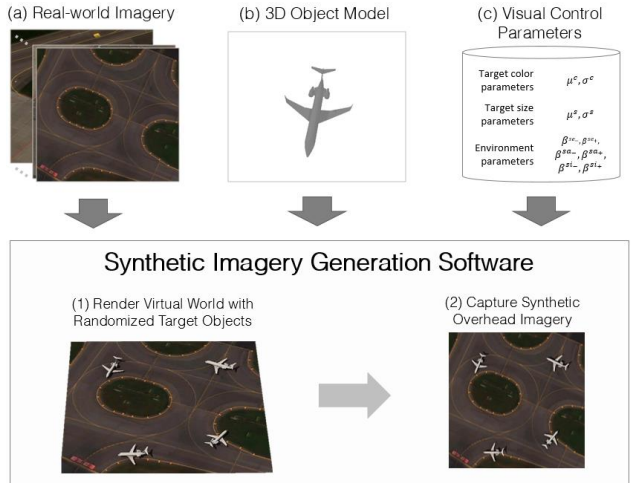

Fig. 1. An illustration of synthetic imagery generation process. Users must provide three primary input: (a) a real-world image which will act as the ground plane in the virtual world, and (b) is a 3D object model, which will be placed randomly on the ground plane. (c) is a set of object characteristics to apply randomly to the 3D object model (e.g. 3D object color, size statistics and environment light). With all user inputs, the synthetic imagery generation software only takes 2 steps to get synthetic output: (1) create virtual world and then (2) capture overhead imagery.)

One major limitation of Synthinel-1, as noted in [19], is that it could not be used to effectively train DNNs alone; it was beneficial only when added to large quantities of real imagery. This makes Synthinel-1 unsuitable for solving zero-shot and few-shot learning problems – an important real-world use-case. The authors attributed this limitation of Synthinel-1 to the visual domain gap of their approach.

Another major limitation of [19] is that there are a large number of free design parameters that must be chosen by users to create synthetic overhead imagery for new problems (e.g., the composition of texture libraries for vegetation, roads and buildings; 3D models for building and trees; road topology; etc.). Furthermore, relatively little insight was provided regarding *how* the authors chose the design of their synthetic imagery, or how the parameters could be chosen for new problems (e.g., new classes of target objects, or new geographic locations).

### B. Contributions of this work

In this work we propose a novel approach for the generation of synthetic overhead imagery that builds upon the process and software in [19] for Synthinel-1, but has the advantages that it (i) yields more photorealistic imagery so that it is effective for zero-shot and few-shot learning; and (ii) the design process of the synthetic imagery is much simpler, and it can be applied to new problems. Our proposed approach is called Synthetic object IMPLantation (SIMPL). In SIMPL we overlay 3-dimensional (3D) target objects onto randomly drawn (2D) patches of real-world overhead imagery, as in Fig. 1.

This general strategy dramatically reduces the complexity of the design process because the scene background – a highly complex component– is already given. The primary design challenge is therefore to choose the properties of the synthetic target objects, such as their colors, shapes, and sizes. In SIMPL we simplify this process by randomly sampling the visual features of individual target objects from probability distributions (e.g., Gaussian) and allowing the designer to set the *parameters* (e.g., mean and variance) of these distributions. Therefore the designers do not need to choose designs for individual targets, and it provides a simple mechanism for them to incorporate their prior knowledge (or uncertainty) about the target objects into the design. We now summarize the major contributions of this work:

1. *SIMPL: a simple approach for generating synthetic overhead imagery for custom object detection tasks.* We detail our data generation approach and release software so that others can replicate it, or adopt it to generate synthetic imagery for *custom* target objects and problems. [2] We also provide guidance for designing SIMPL imagery for custom problems, and validate it in our experiments.

2. *Experiments demonstrating the effectiveness of SIMPL for zero-shot and few-shot learning.* We present experiments demonstrating the benefits of SIMPL for solving zero-shot ($N=0$) and few-shot ($0 < N \leq 150$) learning problems. Our experiments involve training and testing object detection models for seven classes of aircraft.

3. *Design insights of synthetic imagery generation.* We study the impact of several key SIMPL target object design parameters, to provide insight for researchers applying SIMPL to new problems.

The remainder of this document is organized as follows: in section 2 we discuss the SIMPL process for generating synthetic overhead imagery; in section 3 we discuss the materials and methods needed for our experiments; in Section 4 and 5 we discuss the design and results of our zero-shot and few-shot learning experiments, respectively; in Section 6 we analyze the sensitivity of SIMPL

---

[2] https://github.com/yangxu351/synthetic_xview_airplanes/

Table 1: The default properties and sampling distributions of the SIMPL synthetic imagery generation process. The design parameters for each property are also presented on the right. "N" refers to a Normal distribution and "U" refers to a Uniform distribution. Bolded symbols refer to vectors.

| ID | Property description | Property range | Property sampling distribution | Design parameters |
|---|---|---|---|---|
| 1 | Solid color of the 3D target object (RGB) | $\boldsymbol{\theta}^c \in [0,255]^3$ | $\boldsymbol{\theta}^c \sim N(\boldsymbol{\mu}^c, (\sigma^c)^2 \boldsymbol{I})$ | $\boldsymbol{\mu}^c, \sigma^c$ |
| 2 | Length and width of the 3D target object (pixels) | $\boldsymbol{\theta}^s \in (0,\infty)^2$ | $\boldsymbol{\theta}^s \sim N(\boldsymbol{\mu}^s, (\sigma^s)^2 \boldsymbol{I})$ | $\boldsymbol{\mu}^s, \sigma^s$ |
| 3 | Solar elevation angle (degrees) | $\theta^{se} \in [0^o, 90^o]$ | $\theta^{se} \sim U[\beta^{se-}, \beta^{se+}]$ | $\beta^{se-}, \beta^{se+}$ |
| 4 | Solar azimuth (degrees) | $\theta^{sa} \in [0, 360]$ | $\theta^{sa} \sim U[\beta^{sa-}, \beta^{sa+}]$ | $\beta^{sa-}, \beta^{sa+}$ |
| 5 | Solar intensity | $\theta^{si} \in [0, \infty)$ | $\theta^{si} \sim U[\beta^{si-}, \beta^{si+}]$ | $\beta^{si-}, \beta^{si+}$ |

effectiveness to several key design parameters; and in Section 7 we discuss our final conclusions.

## II. THE SIMPL DATA GENERATION PROCESS

In this section, we begin in Section II.A by describing the SIMPL method for generating synthetic overhead imagery, and how the user's input is utilized in each step. In Section II.B we then provide guidance for choosing the input (i.e., the design) of SIMPL data for a custom problem (e.g., target object). Unless otherwise mentioned, we follow these guidelines when designing our experimental synthetic dataset in Section III.C to validate these guidelines. In Section II.C and II.D we discuss the software implementation of the SIMPL method, and its generation speed.

*A. The synthetic data generation process*

Conceptually, SIMPL relies upon overlaying 3-dimensional (3D) object models onto real-world overhead imagery in a virtual envirionement. This process and its major user inputs (i.e., design parameters) are illustrated in Fig. 1. We now discuss each of the major steps of the generation process, and how it makes use of user input.

*Ground plane creation from 2D imagery.* In the first step we create a flat ground plane in the virtual world, and set its appearance to match a user-input overhead image (see Fig. 1(a)). One virtual world will be created for each real-world input image provided by the user, until all input images have been utilized.

*Overlay 3D synthetic objects.* In this step 3D synthetic objects are densely overlaid onto the ground plane of the virtual world. The 3D objects are placed at random spatial intervals and orientations on each overhead image, averaging 120 targets per square kilometer, although this can be easily adjusted in practice. We chose this density for our experiments to create a large number of target instances while minimizing the probability that they will overlap. Each 3D object is a copy of one (or possibly more) 3D object models provided by the user as input (see Fig. 1(b)). Many 3D models can be obtained freely, or for purchase, from online repositories (e.g., CGTrader[3] and Free3D[4]). For example, all of the 3D models used in our experiments in Section 3 were obtained freely from online repositories. Alternatively, if no appropriate 3D models can be found, they can also be designed using CAD software.

*Setting the 3D object visual properties.* We set two visual properties of each emplaced target object: its color ($\boldsymbol{\theta}^c$) and size ($\boldsymbol{\theta}^s$). The values of these two properties for each individual target are randomly drawn from a sampling distribution (e.g., Normal, Uniform). In Table 1 we list these two visual properties, their range of valid values, and their sampling distributions. The sampling distributions listed in Table 1 represent the default distributions employed in our SIMPL software, however, other sampling distributions could be used. By using sampling distributions to set the visual properties of each target, the designer need not manually set the properties of each target. The controllable design parameters for each distribution are presented in Table 1(last column).

*Setting the virtual lighting conditions.* The shadows of 3D objects have been found historically to be important cues for detection (e.g., building detection [23]). Therefore, in the virtual world we set three different lighting properties for *each* real-world input-image: the solar elevation angle ($\theta^{se}$), the solar azimuth angle ($\theta^{sa}$), and the solar intensity ($\theta^{si}$). By varying these parameters it is possible to vary the intensity, length, and orientation of the target shadows; as well as vary lighting-induced textures on the targets themselves, contributing to their photorealism. Like the properties of the the target objects, the particular property values assigned to each image are randomly drawn from sampling distributions, which are presented in Table I.

*Capture synthetic overhead imagery and ground truth bounding boxes.* In the last step of the process we manipulate a virtual camera to take photographs of the virtual world at a desired ground sampling distance (GSD). The height and resolution of the camera are adjusted to achieve a GSD of 0.3 meters per pixel, although this can be adjusted programmatically if needed. To obtain the bounding box annotations of each target object, we set the color of the ground plane and the targets to be white and black, respectively. Then we can (i) segment the targets from the background using a threshold, (ii) identify individual targets by finding connected components, and (iii) then compute the smallest bounding box that encloses each target object.

---
[3] https://www.cgtrader.com/free-3d-models
[4] https://free3d.com/

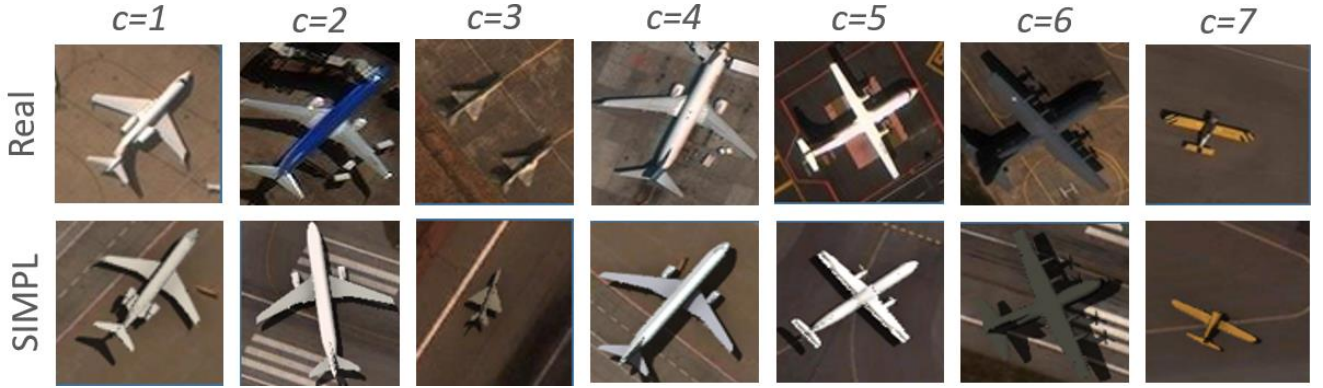

Fig. 2. (top row) Examples of each target class, denoted *c*, considered in our experiments. (bottom row) Examples of SIMPL synthetic imagery created for each target class. We now describe the 3D model used for each class: (c=1) Twin rear enging commercial aircraft (e.g., Boeing 727); (c=2) Wing-mounted twin engine commercial aircraft (e.g., Airbus A320/A330; Boeing 777); (c=3) MiG-21; (c=4) A subset of $c = 2$ that have lower variance in size, shape and colors (e.g., gray/white Boeing 737 or A320); (c=5) Dash 8 Q-Series; (c=6) General Dynamics F-111C; (c=7) Single engine, single seat, monoplanes aircraft (e.g., Supermarine Spitfire ). Additional examples of synthetic imagery can be found in the supplementary materials.

### B. Designing SIMPL imagery for custom problems

In this section we propose guidelines to choose the SIMPL input (i.e., design) for a custom problem (e.g.,. a new target class, or test imagery). To demonstrate the effectiveness of these design procedures, we employed them in the design of our experimental synthetic airplane datasets (see Section III.C). Examples of synthetic overhead imagery patches generated by SIMPL can be found in our supplemental materials.

*Target design parameters.* Here we discuss the following parameters: the 3D object model, the mean color and size ($\boldsymbol{\mu}^c$ and $\boldsymbol{\mu}^s$), and their standard deviations ($\sigma^c$ and $\sigma^s$). To effectively choose these parameters it is assumed that we have *some* knowledge of the target's intrinsic shape, size, and color. This information may come from a few instances of the target object that are available in overhead imagery, or in another imaging modality (e.g., ground level photos). Alternatively, many classes of real-world objects have relatively well-defined visual properties (e.g., cars, airplanes, helicopters), which may be sufficient.

A major theoretical advantage of SIMPL is that we need not know the precise properties of the target objects because we are only choosing the parameters of their *sampling distributions* (e.g., Table 1). Therefore, even if we set the parameters imperfectly (e.g., $\boldsymbol{\mu}^s$ and $\boldsymbol{\mu}^c$) SIMPL is still likely to sample some target objects that are visually similar to the target objects' correct values.

With this in mind, we recommend choosing a 3D model that best reflects the designers knowledge of the average target shape. However, additional 3D models that are variations of the first model can be added to account for uncertainty. The 3D models for our experimental airplane dataset were chosen on this basis, without knowledge of the true ground truth shape of the target aircraft, or even exact knowledge of the aircraft models (see Section III.C).

Similarly, we can choose $\boldsymbol{\mu}^c$ and $\boldsymbol{\mu}^s$ to reflect our best guess about the average color and size of the target objects, and we can increase $\sigma^c$ and $\sigma^s$ to reflect greater uncertainty. To obtain meaningful and replicable results with our experimental SIMPL dataset (see Section III.C), we decided to choose these parameters based upon statistics of our testing imagery, which may not always be possible in practice (e.g., zero-shot learning, where $N = 0$). However, in Section VI we evaluate the sensitivity of SIMPL imagery to the accuracy of some of these parameters to provide insight to designers.

*Non-target design parameters*. Here we discuss the following parameters: the input imagery (Fig. 1(a)), and the solar parameters (ID 3-5 in Table 1). We propose that the input imagery is sampled from the unlabeled test imagery, thereby increasing the similarity between the SIMPL imagery and the testing imagery. This is the approach we used to design our experimental dataset in Section III.C.

To estimate the solar parameters in in Table 1 we propose to inspect and estimate the solar properties (i.e., $\theta^{se}$, $\theta^{sa}$, $\theta^{si}$) for a small a small random sample of the test imagery. Then we find the minimum and maximum values among this sample, and these will become the solar visual design parameters. This is the approach we adopted to choose the solar parameters of our experimental dataset in Section III.C.

*C. The SIMPL software implementation*

With this publication we also release a replicable implementation of SIMPL so that others can utilize it to solve novel problems (e.g., custom target object classes), or build upon the synthetic imagery generation process itself. The core software for SIMPL was built in Python and is therefore open-source, however, it does have a dependency on ESRI's CityEngine, which is a proprietary closed-source software. CityEngine makes it easy to rapidly generate the large-scale virtual worlds, which are used to create SIMPL's synthetic overhead imagery, as well as automatically capture and save both (i) RGB and (ii) ground truth synthetic overhead imagery (via the open-source Python software developed in [19]). Although CityEngine is closed-source it is often readily accessible in industry and academia, as noted in [19]. For example, many Universities have site licenses so that it is freely accessible to researchers. Additional details about our software, and its use, can be found on our online repository.

*D. Speed of the synthetic data generation*

The SIMPL process requires about 40.6 seconds to generate one square kilometer of synthetic RGB imagery and its corresponding ground truth labels. The generated imagery is 0.3m per pixel resolution, so that it matches our real-world experimental datasets. This speed is affected by a variety of factors, but the two most important are (i) the computer's CPU, and (ii) the number of target objects per square kilometer in the imagery. This is because the target objects are 3D, and each one requires a substantial amount of computation (relative to the ground plane) to render; and the CPU is primarily responsible for rendering the scene in our implementation. For our benchmark timing we included 120 target objects per square kilometer (i.e., the value used in all experiments in this work) and we generated the SIMPL data on a Windows 10 operating system with an Intel(R) Core(TM) i9-7920X CPU@2.90 GHz.

## III. MATERIALS AND METHODS

*A. Selection of target objects for experimentation*

In this work we conduct zero-shot and few-shot object detection experiments for a set of seven different classes of aircraft, denoted $C = \{c\}_{c=1}^{7}$, and illustrated in Fig. 2. We chose these target objects for several reasons. First and foremost, aircraft have well-defined visual features (e.g., shape, size and color), with limited intra-class variance. These properties reduce the design complexity of the synthetic data, which allowed us to experiment with more classes and thereby obtain more general experimental results. This also made it easier to perform analyses in Section VI.A, where we control the relative similarity between the synthetic and real imagery – this is much more difficult if the target objects have poorly-defined visual features, with high intra-class variance. The second reason we chose these aircraft is because they were present in an existing publicly-available overhead imagery dataset: the xView dataset [1]. There were also enough unique instances of each class to support experimentation.

*B. The real-world aircraft imagery dataset*

Our experimental dataset of real-world imagery is a subset of the imagery in the xView dataset[1]. xView is a public object detection dataset of satellite imagery that includes imagery from WorldView-3 satellites with 0.3m ground sample distance, and contains 60 target classes. As described, our experiments here focus on the detection of seven different classes of aircraft, however, all aircraft in xView have the same label. Therefore we manually identified and labeled all aircraft that belonged to one of our seven categories. Throughout the paper we will use $c \in \{1, ..., 7\}$ to denote the seven target classes of aircraft, $c = 0$ to denote the "Background" class.

We built our experimental dataset by first splitting the xView imagery into smaller ($608 \times 608$) patches that were small enough to process individually. Then we created seven datasets, denoted $X_c$, each one comprising all patches in xView that contained one or more instances of class $c$. We created $X_0$ by randomly sampling 2,281 patches of imagery from xView that did not contain any instances of the seven aircraft classes.

We split each $X_c$ into two *disjoint* subsets: a training(Tr) dataset, $X_c^{Tr}$; and a testing (Te) datasets, $X_c^{Te}$. The percentage of $X_c$ assigned to each subset depended upon $c$, as illustrated in Fig. 3. For $c \in \{3, ..., 7\}$ we had relatively few instances of target objects, and therefore we assigned all of the data to $X_c^{Te}$; these datasets will only be used for zero-shot learning experiments. The remainder of the datasets were split 80% into $X_c^{Tr}$ and 20% into $X_c^{Te}$, to support both training and testing of models. In order to create more variability in our testing datasets, and have more similar numbers of target instances for each class (this makes $AP_{50}$ metrics more comparable), we augmented $X_c^{Te}$, $c \in C\backslash\{0,2\}$ by rotating each test patch by $\theta \in \{90°, 180°, 270°\}$. We did not augment $X_2^{Te}$ because it already had a relatively large testing set. The final statistics of each datset are shown in Table 2.

Table 2: Summary of our xView aircraft dataset.

| Object class, $c$ | Training dataset, $X_c^{Tr}$ | | Testing dataset, $X_c^{Te}$ | |
|---|---|---|---|---|
| | No. of patches | No. of target instances | No. of patches | No. of target instances |
| 1 | 29 | 50 | 28 | 57 |
| 2 | 107 | 151 | 44 | 91 |
| 3 | 0 | 0 | 12 | 24 |
| 4 | 0 | 0 | 28 | 36 |
| 5 | 0 | 0 | 32 | 40 |
| 6 | 0 | 0 | 12 | 12 |
| 7 | 0 | 0 | 8 | 40 |
| 0 | 1824 | 0 | 456 | 0 |

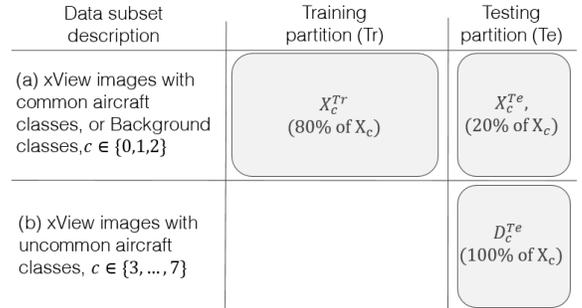

Fig. 3. An illustration of the organization of the experimental datasets for this work.

*C. The SIMPL synthetic airplane dataset*

In this section we summarize how we used the SIMPL approach described in Section II to generate the synthetic training imagery for each target class, denoted $\overline{X}_c^{Tr}$, for our experiments.

*(a) Background imagery:* Following the guidelines proposed in Section II.B, for all $\overline{X}_c^{Tr}$ we randomly sample a set of 450 background patches from $X_0^{Tr}$. Although we could generate an unlimited quantity of synthetic imagery with SIMPL, we found that a dataset size of 450 patches provided a good tradeoff between the time required for training and the performance advantages of using more training data (see Section VI.B for analysis).

*(b) 3D target object models:* All of our target object models were obtained freely from the GCTrader or Free3D website and we chose our 3D models based upon their visual similarity to the target objects in the imagery. Although this strategy cannot always be adopted in practice, SIMPL assumes that the designer has *some* knowledge about the shape of the target objects (see Section II.B). Even though we inspected the test imagery, we had no access to ground truth shape information, nor to the models of the aircraft in the dataset. Therefore, we had imperfect information about the aircraft, in similar fashion to a real-world design scenario. Examples of SIMPL and real-world are presented in Fig. 2, where it is apparent that our 3D models do not perfectly match the real targets.

*(c) Solar parameters:* Following the procedure proposed in Section II.B, we visually inspected a random sample of image patches (approximately 1%), from which we estimated the lower and upper bound of the Uniform distributions for the solar parameters in Table 1.

*(d) Target object parameters.* To obtain $\mu^c$ and $\mu^s$ for each target class, we randomly selected five real-world target objects and measured their average color (RGB values) and size (length and width), respectively. For the RGB averages, we sampled a few individual pixels (e.g., approximately 5) by hand that were located on the target objects. To choose $\sigma^c$ and $\sigma^s$ we created four synthetic datasets for each target class, where each dataset was given a different setting of these parameters. Then we chose the setting that achieved the best performance on the testing dataset for that class. We report the full results of this optimization, and the final optimized parameters for each target class, in our supplemental materials.

Here we designed $(\mu^s, \sigma^s)$ and $(\mu^c, \sigma^c)$ by measuring *some* real-world targets in the test imagery, which may always be adopted in practice (e.g., zero-shot learning, when no target instances are available). This approach was adopted to ensure that our experimental synthetic dataset was relatively similar to the real-world imagery, which is a necessary condition to study the effectiveness of SIMPL, and make our experimental design precise and replicable. To mitigate this limitation for real-world deployment of SIMPL, and to provide insight to designers, in Section VI we evaluate the sensitivity of SIMPL's performance to the accuracy of $\mu^c$ and $\mu^s$ – two key design parameters.

*D. Deep object detector, and training*

For our experiments we use the YoloV3 DNN model [24] for object detection. The YoloV3 is a widely-used single-stage object detection model that has been applied in numerous studies, including overhead imagery [25], [26]. We trained all models with a base learning rate of 0.00579 for fixed number of mini-batch iterations: 9,900. We used a decay rate of 0.1 and a batch size of 8, and an input image size of $608 \times 608$.

In some of our experiments we will use mixed-batch training [9] to utilize training data from one (or more) different populations of training data (e.g., synthetic versus real [19]), while controlling the relative contribution of each training dataset to backpropagation. This is done by constraining each mini-batch to have a fixed number of training samples from each of the training populations. Whenever we use mixed-batch training in this paper, we train for 19,800 iterations (rather than a specified number of epochs) because the dataset sizes are typically larger, and we found more iterations were required to achieve convergence for all models.

*E. Performance metrics*

In this work we use two scoring metrics. The first is the average precision for an intersection-over-union of 0.5 ($AP_{50}$). The $AP_{50}$ metric is widely-used for object detectors in overhead imagery, and therefore we employ it here as well [1], [19], [25]. $AP_{50}$ measures the *average* of the precision of the detector (i.e., the ratio of true detections to false detections) across all values of Recall. This is a useful summary of the performance of a detector, which can be used to compare the overall performance of different detection models.

To complement the $AP_{50}$ metric, we also measure the model's Recall at specific false alarm rates, denoted $R(\alpha)$, where $\alpha$ indicates the false alarms *per square kilometer* of imagery. The $R(\alpha)$ metric is useful because it provides a straightforward mechanism to assess the effort required to use the detector in practical scenarios. In practice, all detections must be manually inspected to remove false detections. A typical detection returned by our detector is approximately 40×40m on average (i.e., 133×133 pixels), which equates to 0.16% of a square kilometer of imagery. Therefore, if our detector returns $\alpha$ false detections per square kilometer, then only $0.16\alpha$% of the total imagery will need to be inspected. In this work, we report $R(1)$ for our detectors due to its simplicity, and because these were sufficient to support our scientific claims. However, we also report $R(0.25)$, $R(0.5)$, and $R(0.75)$ in the supplementary materials**.**

All performance metrics reported in Section 4 and 5 are result of training three models, and then *averaging* the metric obtained for each of the models on the test set. This yields less noisy results, and more apparent trends.

## IV. ZERO-SHOT DETECTION EXPERIMENTS

In these experiments we evaluate whether SIMPL data is effective for zero-shot object detection scenarios, where it is assumed that we do not have any real-world target imagery. With this in mind, we aim to address two main questions with our experiments:
(i) *Can SIMPL substantially reduce the effort and expense needed to manually identify target objects?* Perhaps the most straightforward strategy to address zero-shot problems is to manually inspect the imagery to find the targets of interest (i.e., do not use a detector). Historically this has been the primary approach to analyze overhead imagery. Therefore, it is essential that SIMPL offers clear advantages over this approach.
(ii) *How much real-world training imagery is required to achieve SIMPL-level detection performance?* The collection of new real-world training imagery is another commonly-employed solution to address zero-shot tasks. The time and cost associated with training data collection varies, and therefore we conduct experiments to compare the precise tradeoffs between SIMPL and this common approach. This will also further support the effectiveness of SIMPL by showing that it is comparable to real-world training imagery.

Next we describe the experimental design that we utilized to address these questions.

*A. Experimental design details*

For these experiments we train a YoloV3 detector solely upon the *synthetic* imagery for each target class, $\bar{X}_c^{Tr}$, and then evaluate its performance on a testing dataset comprising real-world target imagery and a large quantity of real-world background imagery (i.e., $X_c^{Te} \cup X_0^{Te}$).

We also include several baseline models for a subset of the classes, $c \in \{1,2\}$, for which we have real-world training imagery. These baseline models allow us to compare the effectiveness of our SIMPL imagery to varying quantities of real-world imagery. We trained detection models on real-world training datasets composed of $X_c^{Tr}(N_i) \cup X_0^{Tr}$, where $N_i$ refers to the number of target instances randomly sampled from $X_c^{Tr}$.

The training data for the baseline models exhibits a severe class imbalance between the target ($X_c^{Tr}$) and non-target ($X_0^{Tr}$) imagery, and therefore we employed mixed-batch training (see Section III.D) to balance the influence of each dataset. For each target class $c \in \{1,2\}$ we optimized the mixed-batch ratio by finding the ratio that achieved the best test performance for the largest value of $N_i$. The full results of this optimization procedure are reported in our supplemental materials. We found that including background (non-target) imagery, and optimizing the mixed-batch training ratio, led to improved performance of our real-world models, providing stronger baselines for our comparisons with SIMPL.

*B. Experimental results*

The results of our zero-shot learning experiments are presented in Table 3. We first note that the maximum value of $N_i$ for class one and two were set differently, based upon the total number of real-world target instances available for that class in our dataset.

First we address experimental question (i), and consider whether SIMPL can reduce the effort needed to detect targets manually. We see that $R(1) \geq 0.5$ for most classes (5 of 7), indicating that over 50% of the target objects are found by the detector, while returning just one false detection per square kilometer (i.e., the false alarm rate when $\alpha = 1$). This is equivalent to one false alarm per 3333×3333 pixel overhead image. Given that the average size (in pixels) of detections is (roughly) 100× 100 pixels, at $\alpha = 1$ our SIMPL-based detectors return just 0.16% of the total test imagery as false detections. These false detections would need to be inspected and removed, however, this represents a significant reduction in the amount of imagery that would otherwise need to be inspected. Therefore, *when no labels are available*, SIMPL-based detectors do provide substantial benefit, enabling the recall of a significant fraction of the objects (over 90% in three of seven classes (4, 5, and 7)), while substantially reducing the quantity of imagery that must be manually inspected.

Table 3: Zero-shot scenario results.

| Class $c$ | Training data | $AP_{50}$ | $R(1)$ |
|---|---|---|---|
| 1 | Synthetic | **0.12** | **0.27** |
|   | Real($N_i$=25) | 0.11 | 0.23 |
|   | Real($N_i$=50) | **0.12** | 0.25 |
| 2 | Synthetic | 0.37 | 0.65 |
|   | Real($N_i$=25) | 0.33 | 0.59 |
|   | Real($N_i$=50) | 0.34 | 0.55 |
|   | Real($N_i$=100) | 0.45 | 0.72 |
|   | Real($N_i$=150) | **0.50** | **0.80** |
| 3 | Synthetic | **0.23** | **0.31** |
| 4 | Synthetic | **0.53** | **1.00** |
| 5 | Synthetic | **0.93** | **0.99** |
| 6 | Synthetic | **0.6** | **0.69** |
| 7 | Synthetic | **0.87** | **0.90** |

Next we consider question (ii), and evaluate how SIMPL compares to real-world training imagery. In Table 3 we report the performance of detectors trained on $c \in \{1,2\}$ with varying levels of real-world target instances for training, $N_i$. As expected, the performance of each target class increases in proportion to $N_i$. We see that SIMPL-trained detectors usually outperform real-world models when $N_i \leq 50$, providing evidence that SIMPL imagery does indeed resemble real-world training imagery. Furthermore, SIMPL imagery is (roughly) equivalent to training on some limited quantity of real-world imagery. As discussed in Section Section I, collecting $N_i \cong 50$ real-world instances can be time-consuming and costly, especially if the target object is rare. Using synthetic data alone yielded performance metrics similar to that from having 50 to 100 real image examples in the training set. Ultimately the superior performance of real-world imagery may outweight the costs of data collection, however, this decision will be application dependent.

## V. FEW-SHOT DETECTION RESULTS

In these experiments we evaluate whether SIMPL data are effective for few-shot object detection scenarios, where it is assumed that we have a small set of real-world target instances ($25 \leq N_i \leq 150$) available for training. In this case we want to improve the performance of our detectors by augmenting the limited real-world training imagery with SIMPL imagery. With this in mind, we aim to address two questions with our experiments:

(i) *Is augmention with SIMPL imagery beneficial when training our object detectors?*

(ii) *Does the benefit of augmentation with SIMPL imagery depend upon the quantity of real-world target instances?* Due to the visual domain gap between real-world and SIMPL imagery, it is likely that SIMPL imagery will become increasingly unhelpful as $N_i$ increases. Therefore it will be useful to estimate how the benefits of SIMPL vary with respect to the number of real training images, $N_i$.

Next we describe the experimental design that we utilized to address these questions.

### A. Experimental design details

For these experiments we first created baseline models by training solely on real-world training imagery. We did this by following the same procedure here as in Section 4, and therefore the baseline results in Section 4 ( Table 3) and here (Table 4) are identical. We then evaluated the benefits of SIMPL for few-shot learning by augmenting the training imagery for each baseline model with SIMPL imagery of the same target class. Specifically, we created training datasets of the form $\bar{X}_c^{Tr} \cup X_c^{Tr}(N_i) \cup X_0^{Tr}$ where $N_i$ refers to the number of real-world training instances, and $\bar{X}_c^{Tr}$ refers to synthetic imagery for class $c$. Similar to the baseline models, we trained each of the models with mixed-batch training using two populations of training imagery: $\bar{X}_c^{Tr} \cup X_c^{Tr}(N_i)$ and $X_0^{Tr}$. Also similar to the baselines, for each target class we optimized the mixed-batch ratio once for the largest value of $N_i$, and then used this ratio for all other experiments on that class. The full results of this optimization procedure are reported in our supplement.

### B. Results

The results of our few-shot learning experiments are presented in Table 4. We begin by addressing scientific question (i), and we find that augmentation with SIMPL imagery nearly always improves the performance of the resulting detection models. Furthermore, these performance improvements are usually substantial (e.g., $\geq 34\%$ $R(1)$ in most cases). These results suggest that SIMPL imagery can be highly beneficial in few-shot detection scenarios.

We next consider scientific question (ii): do the benefits of SIMPL vary as a function of $N_i$? We see that for both target classes, the performance benefits of SIMPL compared to the baseline models gradually decrease as $N_i$ increases. These results therefore suggest that SIMPL does indeed decrease in effectiveness as more real-world training imagery becomes available. Importantly, for the values of $N_i$ that we consider, there appears to be relatively little risk of using SIMPL, since it only decreased performance in one case, for $N_i = 150$: the largest value of $N_i$ that we consider. Furthermore, the performance degradation in that case was relatively modest ($\leq 2\%$ for both $AP_{50}$ and $R(1)$). These results further suggest that SIMPL is effective for few-shot detection scenarios.

Table 4. Results of few-shot scenario experiments. For convenience, we also report the percentage change in $R(1)$ when adding synthetic imagery in training, denoted %Δ$R$.

| Class $c$ | No. real instances | Add Synthetic? | $AP_{50}$ | $R(1)$ | %Δ$R$ |
|---|---|---|---|---|---|
| 1 | $N_i$=25 | No | 0.11 | 0.23 | +73 |
| | | Yes | **0.19** | **0.40** | |
| | $N_i$=50 | No | 0.12 | 0.25 | +63 |
| | | Yes | **0.20** | **0.37** | |
| 2 | $N_i$=25 | No | 0.33 | 0.59 | +34 |
| | | Yes | **0.42** | **0.79** | |
| | $N_i$=50 | No | 0.34 | 0.55 | +44 |
| | | Yes | **0.45** | **0.79** | |
| | $N_i$=100 | No | 0.45 | 0.72 | +18 |
| | | Yes | **0.47** | **0.82** | |
| | $N_i$=150 | No | **0.50** | **0.80** | -1 |
| | | Yes | 0.49 | 0.79 | |

## VI. SIMPL DESIGN AND SENSITIVITY ANALYSIS

In this section we analyze the impact of several key design parameters on the effectivenss of SIMPL imagery. This provides insight for users regarding the design of SIMPL imagery for novel problems.

### A. Impact of target color and size accuracy

For our synthetic airplane dataset, we directly measured the colors and sizes of the target aircraft in the imagery to obtain $\boldsymbol{\mu}^c$ and $\boldsymbol{\mu}^s$ (see Section III.C for details). Assuming that these estimates are relatively accurate, we can vary their accuracy by adding progressively larger offsets to them. Therefore, to study the impact of inaccuracy on the effectiveness of SIMPL, in Fig. 4 we present the percentage *change* in $AP_{50}$ of our SIMPL-trained detectors as a function of percentage offset (i.e., error) in $\boldsymbol{\mu}^c$ and $\boldsymbol{\mu}^s$. These models were all evaluated on the same testing datasets used in our zero-shot and few-shot experiments in Sections IV and V.

The results for $\boldsymbol{\mu}^c$ in Fig. 4 (a) indicate that errors in color generally have a limited impact on performance. The average performance among the different classes is largely unchanged as the error in $\boldsymbol{\mu}^c$ increases, although the variance in performance does increase. The performance of class six drops substantially in some cases, however, the effect size of errors in $\boldsymbol{\mu}^c$ is still unclear because of the high variance in the performance of class six.

By contrast, the results for $\boldsymbol{\mu}^s$ in Fig. 4(b) indicate that errors in size can have a substantial and negative impact on performance, although the impact does vary substantially depending upon the class. For example, for $c \in \{1,2,7\}$, errors in $\boldsymbol{\mu}^s$ result in litte (and possibly positive) impact, while for other classes we see substantial degradations.

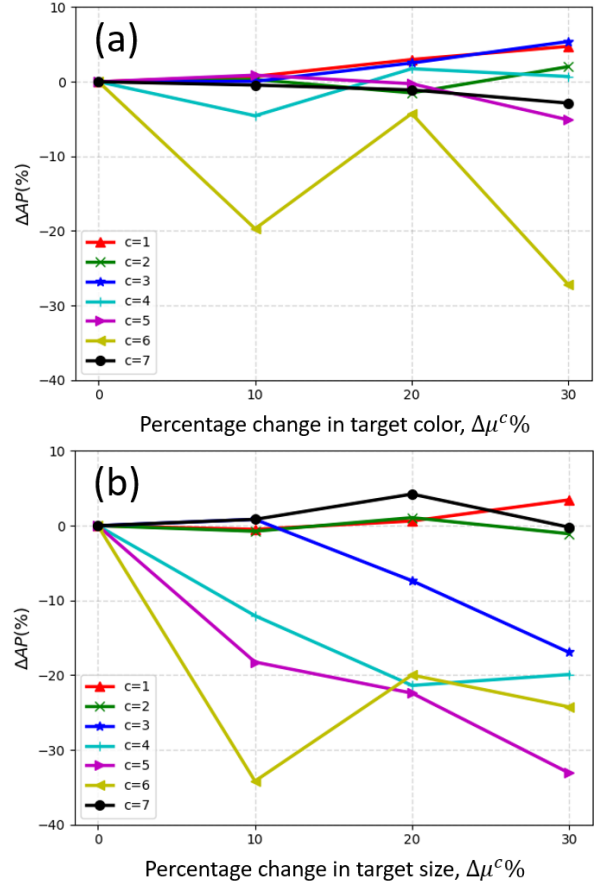

Fig. 4. (a) The percentage change in $AP_{50}$ as a function of the percentage change in $\mu^c$. (b) The percentage change in $AP_{50}$ as a function of the percentage change in $\mu^s$.

Therefore, the results here indicate that it is much more important to choose an accurate size than an accurate color.

### B. Quantity of synthetic training imagery

Here we evaluate the changes in performance of SIMPL-trained detectors as we vary the quantity of training data, given in terms of the number of synthetic patches, $K$. The results of this experiment are reported in Fig. 5 for each of the seven classes. As expected, increasing $K$ generally leads to improved performance, with diminishing returns as $K$ grows larger. One exception is class six, however this appears to be due to the high overall variance in its performance rather than a clear negative trend, similar to its behavior in Section VI.A. Class three indicates a modest downward trend in performance as $K$ increases. We hypothesize that this arises because the initial sample of synthetic training imagery – when when $K \in \{28, 56\}$ – was especially well-suited for the testing imagery providing an initial positive performance bias that is diluted as more synthetic imagery is added to the initial set (as $K$ grows).

In our zero-shot and few-shot experiments we utilized $K = 450$ because this provided sufficiently good performance to corroborate the effectiveness of SIMPL however, based upon these results, better performance may be achievable by generating and utilizing more synthetic training imagery, at the cost of longer training time.

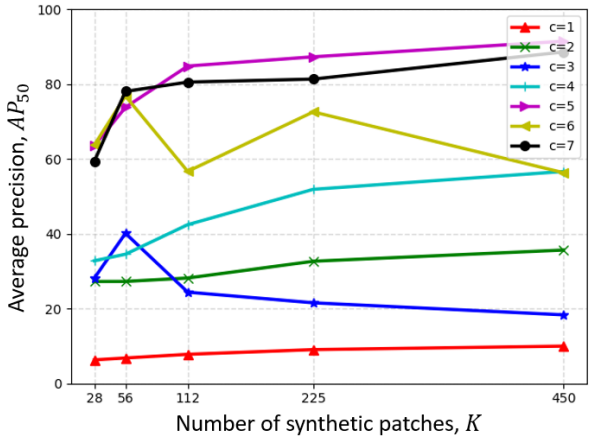

Fig. 5. The $AP_{50}$ metric of SIMPL-trained detectors as a function of the quantity of SIMPL data that was used to train them, $N$.

## VII. CONCLUSIONS

In this work we investigated the use of *synthetic* overhead imagery as a general approach to address zero-shot and few-shot object detection problems. We proposed a novel approach, called SIMPL, that provides a simple approach to design and generate synthetic overhead imagery that can be used to train object detection (e.g., deep neural networks (DNNs)), either alone (i.e., zero-shot settings), or in combination with limited quantities of real-world imagery (i.e., few-shot settings). We conducted experiments for the detection of seven classes of real-world aircraft in the publicly-available xView benchmark dataset. From our experiments we can draw several conclusions:

- *SIMPL is effective for zero-shot learning.* We found that the SIMPL-trained detectors in our experiments usually detect most targets, while yielding just one false detection per square kilometer – a substantial reduction in the quantity of imagery that must be inspected. We also found that SIMPL-trained detectors achieve performance comparable to detectors trained with $50 \leq N_i \leq 100$ real-world target instances.
- *SIMPL is effective for few-shot learning.* We found that augmentation of training imagery with SIMPL generally improved performance of trained detectors when $25 \leq N_i < 100$ real-world instances were available. When $N_i \leq 50$, the improvements were substantial (e.g., $\geq 20\%$ increase in average precision and recall).
- *SIMPL is the first approach for generating effective synthetic overhead imagery for custom object detection problems.* Although recent work has investigated synthetic overhead imagery, SIMPL is the first approach to offer a precise and simple set of user inputs, as well as propose a practical procedure to design data for custom detection problems. We validate SIMPL in our zero-shot and few-shot experiments, and we also provide sensitivity analysis for some key design parameters.

In general the results here demonstrate the potential value of synthetic overhead imagery for training object detectors, and provides the resources for researchers and practitioners to deploy synthetic imagery for new problems, develop more effective design procedures, or improve upon the synthetic data generation process in the future.

## ACKNOWLEDGMENTS

We thank the Duke University Energy Initiative for their support for this work.